\documentclass[11pt]{article}

\usepackage{chngcntr} 
\usepackage{amsmath}
\usepackage{enumerate}
\usepackage{fancyhdr}
\usepackage{color}
\usepackage{listings}
\usepackage{graphicx}
\usepackage{amssymb}
\usepackage{mathtools}

\usepackage{hyperref}
\usepackage{booktabs}

\usepackage[numbers]{natbib}
\usepackage{fullpage}

\usepackage{algorithm}
\usepackage{amsmath,amsthm,amsfonts}
\usepackage{algorithmic}
\usepackage{graphicx}
\usepackage{textcomp}
\usepackage{microtype}

\usepackage{amsmath,amsthm}

\providecommand{\abs}[1]{\left\lvert#1\right\rvert}
\providecommand{\norm}[1]{\left\lvert\left\lvert #1\right\rvert\right\rvert}

  \providecommand{\R}{\mathbb{R}} % Reals
  \DeclareMathOperator{\E}{{\mathbb E}}
  \providecommand{\E}[1]{{\mathbb E}\left.#1\right. }     %expectation
  \DeclareMathOperator*{\argmin}{arg\,min}

\newtheorem{theorem}{Theorem}
\newtheorem{proposition}{Proposition}
\newtheorem{lemma}[theorem]{Lemma}

\newtheorem{definition}{Definition}
\newtheorem{remark}{Remark}
\newtheorem{assumption}{Assumption}

\definecolor{mydarkblue}{rgb}{0,0.08,0.45}
\renewcommand{\cite}[1]{\citep{#1}}
\newcommand{\I}[1]{\mathbb{I}\{#1\}}
\newcommand{\ip}[2]{\langle #1 , #2 \rangle}
\usepackage{makecell}
\usepackage{subfig}

\definecolor{lightgray}{gray}{0.5}

\usepackage[papersize={8.5in,11in},top=1in,bottom=0.75in,left=0.75in,right=0.75in]{geometry}
\usepackage[colorinlistoftodos]{todonotes}
\usepackage{enumitem}
\usepackage{tablefootnote}

\begin{document}

\title{\textbf{Client Selection in Federated Learning with Data Heterogeneity and Network Latencies}}
\author{Harshvardhan, Xiaofan Yu, Tajana \u Simuni\'c Rosing, Arya Mazumdar  \vspace{2mm} \\
Computer Science and Engineering and Hal\i c\i o\u glu Data Science Institute\\
University of California San Diego\\
\vspace{1.5mm}
\{hharshvardhan, x1yu, tajana, arya\}$@$ucsd.edu
}
\maketitle

\begin{abstract}
Federated learning (FL) is a distributed machine learning paradigm where multiple clients conduct local training based on their private data, then the updated models are sent to a central server for global aggregation. The practical convergence of FL is challenged by multiple factors, with the primary hurdle being the heterogeneity among clients. This heterogeneity manifests as \textit{data heterogeneity} concerning local data distribution and \textit{latency heterogeneity} during model transmission to the server. While prior research has introduced various efficient client selection methods to alleviate the negative impacts of either of these heterogeneities individually, efficient methods to handle real-world settings where both these heterogeneities exist simultaneously do not exist. In this paper, we propose two novel theoretically optimal client selection schemes that can handle both these heterogeneities. Our methods involve solving simple optimization problems every round obtained by minimizing the theoretical runtime to convergence. Empirical evaluations on 9 datasets with non-iid data distributions, $2$ practical delay distributions, and non-convex neural network models demonstrate that our algorithms are at least competitive to and at most $20$ times better than best existing baselines.

\end{abstract}


\maketitle
\section{Introduction}
\label{sec:intro}
As edge devices (sensor devices, mobile phones) become more ubiquitous and computationally powerful, frameworks that can utilize their decentralized computation have gained significant interest in recent years. 
Federated Learning (FL)~\cite{pmlr-v54-mcmahan17a,konevcny2016federated} is one such decentralized Machine Learning framework, which trains a single model collaboratively on distributed edge devices (referred to as \textit{clients}) using local data and local computational resources.
In a single round of the typical FL algorithm, FedAvg~\cite{pmlr-v54-mcmahan17a}, a central server distributes a global model to selected clients. Each client performs local training using its private data and sends the trained model back to the server.
Finally, the server aggregates the trained models by averaging the weights of models from different clients. FL has been widely used for a variety of applications in Internet of Things (IoT), including predictive models in smartphones~\cite {gboard,mobile_2}, healthcare~\cite{Xu2020,healthcare_2}, and autonomous driving~\cite{autonomous_driving}. 

\begin{figure}[!tbp]
\centering
\setlength{\tabcolsep}{0.2pt}
\begin{tabular}{cccc}    
    \includegraphics[width=0.4\textwidth]{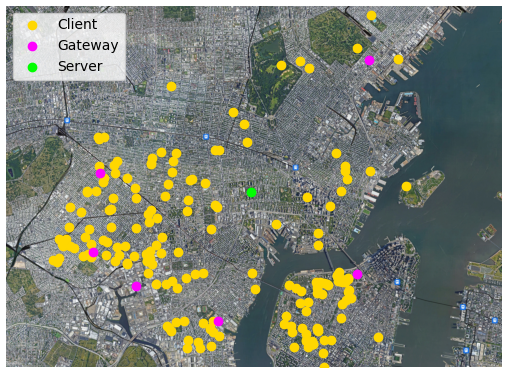} 
    &\hfill
    \includegraphics[width=0.4\textwidth]{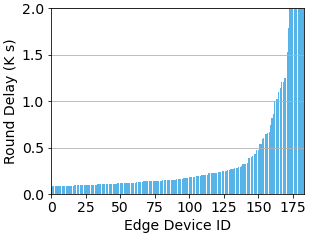}
\end{tabular}
\caption{\small Left: NYCMesh topology. Right: The round delay distribution of all edge devices.}
\label{fig:nycmesh}
\end{figure}

However, disparities persist in the practical and large-scale deployment of FL in real-world scenarios, due to inherent heterogeneities of various types.
One prevalent challenge is \textit{data heterogeneity} in FL. Given that data are generated locally on edge devices, these devices may exhibit non-iid data distributions. This dissimilarity can lead to slow convergence or, in extreme cases, result in a final model which has subpar performance and whose losses diverge~\cite{balakrishnan2022diverse,pmlr-v151-jee-cho22a}.
Furthermore, within a real-world network, different edge devices enjoy varying computational power, distance and network bandwidth in their connection to the server~\cite{nishio2019client,yu2023async}.
All of these system heterogeneities impact the delays and performance in FL, and can be reflected as diverse \textit{round delays} among clients, which is defined as the time taken to perform local computation and transmit the trained model to the server.
NYCMesh~\cite{nycmesh}, as shown in Fig.~\ref{fig:nycmesh} (left), acts as an example of a wireless mesh network setup in New York City. We treat the supernode as the server, and the non-hub nodes as local edge devices. We then simulate round latencies using the urban propagation model in ns-3~\cite{ns3} and present the distribution in Fig.~\ref{fig:nycmesh} (right). A clear long-tail pattern can be observed, with 10\% of the clients taking more than 1000 seconds to return their updates, which can be attributed to the uncertain wireless propagation environment in a city.
This substantial \textit{latency heterogeneity} may lead to unacceptable delays in conventional FedAvg, where the server waits for the slowest selected client to return before aggregation occurs. In summary, training high-quality models within a short timeframe poses a significant challenge in real-world FL use cases.

One natural solution to this problem is to select a subset of devices for training, which is known as client selection in FL. 
The convergence of the FL algorithms under wall-clock time is jointly influenced by the data heterogeneity and round delays within the chosen subset.
Most client selection works handle data and latency heterogeneity separately -- ~\cite{pmlr-v151-jee-cho22a,balakrishnan2022diverse,pmlr-v139-fraboni21a,pmlr-v202-zhang23y,chen2022optimal} to accommodate data heterogeneity and ~\cite{reisizadeh_straggler-resilient_2020} to address latency heterogeneity. 
Several recent studies~\cite{nishio2019client,Lai2020OortEF,delay_paper_2,fast_convergent} jointly considered data and latency heterogeneities. While these studies have introduced various heuristic approaches, they fall short of providing a rigorous theoretical analysis.
To the best of our knowledge, only ~\cite{luo_tackling_2022, lat_het}  optimize both data and latency heterogeneity in client sampling for FL, accompanied by theoretical guarantees. We use ApproxOptSampling to refer to the method of ~\cite{lat_het}. However, both their methods use a loose modeling for data heterogeneity, which fails to distinguish between clients having data heterogeneity for even simple use-cases like linear regression. Additionally, these methods also require non-trivial and computationally expensive warm-up strategies which do not scale well with the number of clients.

To obtain a \textit{theoretically sound} and \textit{computationally efficient} client selection scheme for tackling both delay and heterogeneity, we use the theoretical convergence analysis of FL~\cite{pmlr-v119-karimireddy20a,Li2020On} for non-iid data in the presence of delays to come up with an appropriate metric to select clients. Further, we show that this metric is easy to optimize, which results in our client selection schemes, \texttt{DelayHetSubmodular} and ~\texttt{DelayHetSampling}, each representing a distinct flow of methods in client selection. We now list our main contributions.

\subsection{Main Contributions}
    \textbf{Algorithm Design} Our proposed schemes, \texttt{DelayHetSubmodular} and \texttt{DelayHetSampling}, in Algorithms~\ref{alg:submodular} and \ref{alg:optimalsampling} respectively, minimize the total runtime to achieve a particular accuracy by balancing the tradeoff between client delays and client data heterogeneity. \texttt{DelayHetSubmodular} selects a specific set of clients, while \texttt{DelayHetSampling} provides a sampling distribution from which these clients can be sampled. We provide an interpretation of our precise characterization of heterogeneity in Assumption~\ref{assump:pair_het} into terms that affect convergence speed (feature heterogeneity) and convergence error(optima heterogeneity) by analyzing linear regresion. We utilize this insight to provide appropriate implementations of our client selection schemes  for even deep NN models. 

    \textbf{Theoretical Optimality} These schemes are derived by estimating the total runtime theoretically for a class of smooth and P\L losses, under arbitrary delays and pairwise gradient heterogeneity (Assumption~\ref{assump:pair_het}). When the overall heterogeneity in the system is bounded (Assumptions~\ref{assump:bdd_het} and~\ref{assump:bdd_het_sampl}), both these schemes are theoretically optimal in minimizing runtime (Remarks~\ref{rem:submodular} and \ref{rem:sampling}).

    \textbf{Empirical Performance} We run our client selection schemes on $9$ datasets and compare them with state-of-the-art client selection baselines in FL. We use a synthetic dataset (Quadratic) with feature heterogeneity. We also use simulated datasets, where we add two different forms of heterogeneity, via rotations and class imbalance (Dirichlet), to MNIST~\cite{deng2012mnist}, FashionMNIST~\cite{xiao2017fashionmnist} and CIFAR10~\cite{cifar10}. Finally, we use real federated datasets, FEMNIST and Shakespeare, from LEAF~\cite{leaf} database. All models except Quadratic use NN models with last-layer inputs as features. Our baselines are random client sampling, DivFL~\cite{balakrishnan2022diverse}, Power-of-Choice~\cite{pmlr-v151-jee-cho22a}), FLANP~\cite{reisizadeh_straggler-resilient_2020}) and  (ApproxOptSampling~\cite{lat_het}). Further, we use two delay distributions -- synthetic delays and NYCMesh delays. In terms of total runtime to converge to a target test accuracy, we outperform all baselines in all datasets, except for certain cases of MNIST and FashionMNIST, where we are competitive with the best baseline (Tables~\ref{tab:uniform_runtime} and \ref{tab:nycmesh_runtime}).

\section{Related Works}
\label{sec:related_works}
%Federated Learning has seen remarkable progress since the first FL algorithm FedAvg~\cite{pmlr-v54-mcmahan17a} was published. Initially, theoretical guarantees could only be provided under no heterogeneity assumptions~\cite{stich2018local}. However, this was extended to partial participation~\cite{pmlr-v119-karimireddy20a,Li2020On}. 
Federated Learning has seen remarkable progress since the introduction of the first FL algorithm, FedAvg~\cite{pmlr-v54-mcmahan17a}. Theoretical guarantees were only attainable under the assumption of no heterogeneity initially~\cite{stich2018local}, but was expanded to non-iid data distribution recently~\cite{pmlr-v119-karimireddy20a, Li2020On}. 
%However, recent progress expanded to encompass partial participation scenarios~\cite{pmlr-v119-karimireddy20a, Li2020On}. 
The only client selection scheme analyzed in these works is random client selection, which performs poorly in the presence of various heterogeneity.

\textbf{Address Data Heterogeneity in FL} Among the works which tried to tackle the heterogeneity problem, ~\cite{chen2022optimal} proposed an unbiased sampling scheme which minimizes sampling variance borrowing techniques from importance sampling~\cite{pmlr-v97-horvath19a}, ~\cite{pmlr-v151-jee-cho22a} propose a biased sampling scheme that samples clients with the largest local losses and ~\cite{fast_convergent} samples clients proportional to their decrease in loss value. ~\cite{balakrishnan2022diverse} improve over these methods by handling the heterogeneity of the complete client set instead of individual clients. They sample a representative subset of clients via submodular maximization~\cite{fujishige2005submodular} using techniques from coreset selection~\cite{pmlr-v119-mirzasoleiman20a}. While these methods address a more general notion of data heterogeneity, for the special case of label imbalance, ~\cite{pmlr-v202-zhang23y} sample clients such that the classes among the selected clients are balanced.  While being fast in convergence (number of rounds), all these schemes suffer in terms of wall-clock time to convergence when the selected clients have large delays.

\textbf{Address Latency Heterogeneity in FL} To tackle slow clients referred to as  stragglers, ~\cite{nishio2019client} select clients based on their computation and communication time. ~\cite{Lai2020OortEF} provide a metric per client based on gradient norms and client delays. Note that both schemes~\cite{nishio2019client,Lai2020OortEF} are heuristics and lack theoretical analysis. ~\cite{reisizadeh_straggler-resilient_2020} provide a theoretically optimal scheme for latency heterogeneity, in which they pick the fastest clients initially and keep adding slower clients incrementally. In terms of wall-clock time to convergence, these methods suffer when the fastest clients have adverse data heterogeneity.

\textbf{Address both latency and data heterogeneity in FL} Two existing works~\cite{luo_tackling_2022,lat_het} attempt to tackle the heterogeneity and delay problem simultaneously. The core-idea behind these is to incorporate delays into the client sampling scheme of ~\cite{chen2022optimal} and minimize the total runtime.
While our delay modeling resembles these papers, these suffer from a poor characterization of data heterogeneity and an expensive warm-up procedure. Their data heterogeneity assumptions cannot distinguish between clients with heterogeneous data, even for simple models like linear regression. Further, their warm-up procedure, which is necessary to run their algorithms, performs a few rounds of full client participation, thus adding a drastic latency overhead. For around $10$ rounds of warm-up, this overhead is $5.5$ hours for NYCMesh from Figure~\ref{fig:nycmesh}. In contrast, our data heterogeneity definitions are tighter (see Assumption~\ref{assump:pair_het}), and our warm-up requires only $1$ round of full client participation, which is incurred for all baselines ~\cite{balakrishnan2022diverse,chen2022optimal,pmlr-v151-jee-cho22a,reisizadeh_straggler-resilient_2020} except random client selection.

In summary, existing works lack theoretical optimal and computationally efficient schemes for client selection addressing both data and latency heterogeneity in FL.

% \textcolor{red}{xxxxxxxx}

%Note that there are several aspects of Federated Learning that we do not cover, for instance privacy, client availability, energy consumption, or better FL optimization algorithms. We direct the reader to ~\cite{fl_survey} for an overview of these aspects. algorithm requires solving several subproblems, and is hard to
\section{Motivation}
\label{sec:motivation}

Before we dive into our client selection schemes, we conduct a motivational study of existing methods in three distinct scenarios: (a) FL with solely data heterogeneity among clients, (b) FL with only network latency heterogeneity among clients, and (c) FL with both data and latency heterogeneity among clients. 
The purpose of the study is to observe how various client selection schemes perform under combinations of data and latency heterogeneities.

\begin{figure*}[!tbp]
\subfloat[\normalsize  Only Data Heterogeneity.]{
\includegraphics[width=0.3\textwidth, height=3.0cm]{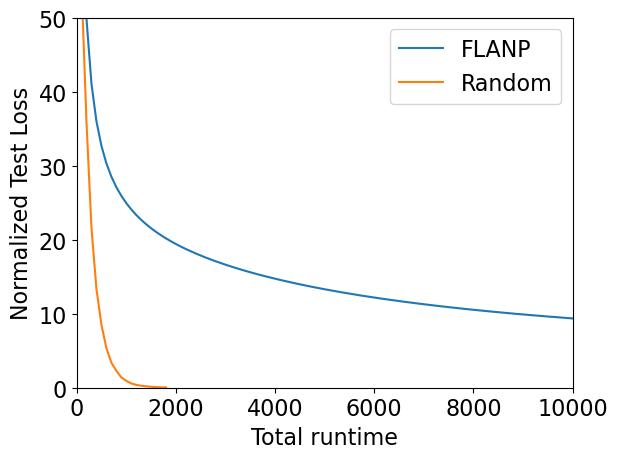}
\label{fig:quadratic_nodelay}
}
\subfloat[\normalsize Only Latency Heterogeneity.]{
\includegraphics[width=0.3\textwidth, height=3.0cm]{figures/quadratic_plot_nodelay.jpg}
\label{fig:quadratic_onlydelay}
}
\subfloat[\normalsize Both Heterogeneities.]{
\includegraphics[width=0.3\textwidth, height=3.0cm]{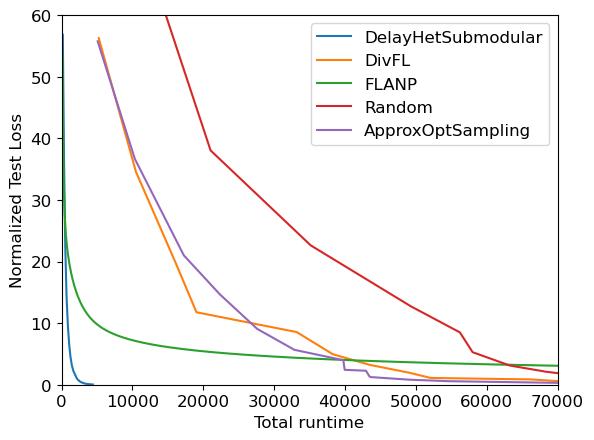}
\label{fig:quadratic_nycmesh}
}
\caption{\normalsize Total wall-clock runtime of different client selection algorithms for Quadratic dataset under three distinct scenarios. 
(a) Performance of random client selection FLANP~\cite{reisizadeh_straggler-resilient_2020}, which handles delays, on pure heterogeneous data without network delays. (b) Performance of random and heterogeneity-aware client selection schemes, DivFL~\cite{balakrishnan2022diverse} and Power-of-Choice~\cite{pmlr-v151-jee-cho22a}, on iid data with NYCMesh latency distribution. (c) Performance of all baselines handling heterogeneous data with heterogeneous network latency distribution from NYCMesh. \texttt{DelayHetSampling} and \texttt{DelayHetSubmodular} are our proposed client selection methods.}
\label{fig:motivation}
\end{figure*}

We experiment using linear regression with $m=100$ clients, where each client possesses $n=100$ training datapoints. Each datapoint is characterized by $d$-dimensional features ($d=500$), denoted as $x$, sampled from a Gaussian distribution, and real-valued labels, denoted as $y$, generated by $\ip{w^\star}{x} + \zeta$. Here, $\zeta$ represents a zero-mean Gaussian noise and $w^\star \in \R^d$ is the optimal model.
%Each client contains $n=100$ training datapoints.  Each datapoint consists of $d$-dimensional($d=500$) features, $x$, sampled from a Gaussian distribution and real-valued labels, $y$, generated according to $\ip{w^\star}{x} + \zeta$, where $\zeta$ is a zero-mean gaussian noise and $w^\star \in \R^d$ is the optimal model. 
We configure the data and latency heterogeneity as follows:
\begin{itemize}
    \item \textit{Data Heterogeneity:} Each client samples its features $x$ from a Gaussian distribution with a different covariance matrix. The covariance matrices of all clients have the same eigenvectors, but their eigenvalues are sampled uniformly from the range $[1,10]$.

    \item \textit{Latency Heterogeneity:} We set the latency of each client based on the NYCMesh~\cite{nycmesh} topology as shown in Fig.~\ref{fig:nycmesh}. Unlike the data center setting, wireless mesh networks such as NYCMesh consider the network latency heterogeneity that is prevalent in practical deployments. Specifically, we extract the installed node locations, including longitude, latitude, and height, from NYCMesh's website. We then simulate the average round latency per client by providing the client's and the server's locations as input to the \texttt{HybridBuildingsPropagationLossModel} in ns-3~\cite{ns3}, a discrete-event network simulator.
    To account for network uncertainties, we introduce a log-normal delay on top of the mean latency at each local training round. The simulated network delays, depicted in Fig.~\ref{fig:nycmesh}, replicate the measurement results presented in~\cite{sui2016characterizing}.
\end{itemize}
%We use this linear regression framework throughout our paper, first to interpret our algorithms in Section~\ref{sec:interpretation} and then to test the empirical performance of our algorithms in Section~\ref{sec:experiments} where it is referred to as Quadratic dataset. Additional details for this setup can be found therein.
In Figure~\ref{fig:motivation}, we report the square loss normalized by $\sqrt{d}$ on each client's test set as the convergence metric, with regard to the total wall-clock runtime, which is computed the sum of the maximum delays of selected clients in each round.
%, for different client selection schemes for the three cases of data heterogeneity, delay heterogeneity and both data and delay heterogeneity. 
We consider existing client selection schemes that only address data heterogeneity (DivFL~\cite{balakrishnan2022diverse} and Power-of-Choice~\cite{pmlr-v151-jee-cho22a}), and those which only address heterogeneity in network latencies (FLANP~\cite{reisizadeh_straggler-resilient_2020}), the random baseline of selecting clients randomly, the ApproxOptSampling baseline from ~\cite{lat_het} with only $1$ warm-up round, and one of our proposed methods, \texttt{DelayHetSubmodular}.

\paragraph{Findings and Key Insights}

From Figure~\ref{fig:quadratic_nodelay}, in the presence of data heterogeneity, FLANP which can handle latency heterogeneity performs worse than even random.
In Figure~\ref{fig:quadratic_onlydelay}, the heterogeneity-aware baselines of DivFL and Power-of-Choice obtain similar performance as random client selection in the absence of data heterogeneity. 
Therefore, these client selection schemes addressing only one of data or latency heterogeneity perform no better than random in the presence of the other. This motivates us to study schemes that can handle data and delay heterogeneity simultaneously. 
From Figure~\ref{fig:quadratic_nycmesh}, even the baseline ApproxOptSampling~\cite{lat_het}, which claims to handle both latency and data heterogeneity is only slightly better than the best baseline. In contrast, one of our proposed methods, \texttt{DelayHetSubmodular}, outperform all baselines by a significant margin. 

\section{Problem Setting}
\label{sec:problem_setting}
\begin{algorithm}[t!]
\begin{algorithmic}
\REQUIRE Step size $\eta$, Initial iterate $w_0$, Number of Local steps $E$, Number of rounds $R$
\ENSURE Final Iterate $w_{R}$
\STATE \underline{\texttt{Server(r):}}
\FOR{$r=0$ to $R-1$}
\STATE $S_r, \{\alpha_{i}\}_{i\in S_r} \gets \texttt{ClientSelection()}$
\STATE Send $w_r$ and receive $w_{r,E}^i$ from each clients $i \in S_r$ 
\STATE $w_{r+1} \gets \sum_{i\in S_r} \alpha_i w_{r,E}^i$
\ENDFOR
\STATE \underline{\texttt{Client($i, w_r$):}}
\STATE Receive $w_r$ from server,  $w_{r,0}^i \gets w_r$
\FOR{$e=0$ to $E-1$}
\STATE $w_{r,e+1}^i \gets w_{r,e} - \eta \nabla f_i(w_{r,e})$
\ENDFOR
\STATE Send $w_{r,E}^i$ to server.
\end{algorithmic}
    \caption{FedAvg using \texttt{ClientSelection}}
\label{alg:fedavg}
\end{algorithm}

In this section, we formally define our problem setting.
 There are $m$ machines each with its loss function $f_i : \R^d \to \R$ and delay $\tau_i$. We need to minimize the average loss across clients defined below.
\begin{align}\label{eq:main_opt}
\min_{w\in \R^d} f(w) \coloneqq \frac{1}{m}\sum_{i=1}^m f_i(w)
\end{align}
We use uniform weights of $\frac{1}{m}$ for all clients in ~\eqref{eq:main_opt}, but our method can also handle weights proportional to the number of data points. Let $w^\star \in \argmin_{w\in \R^d} f(w)$ be one of the minimizers. Then, our goal is to obtain a $w\in \R^d$ that is $\epsilon$-minimizer in loss value, i.e., $f(w) - f(w^\star) \leq \epsilon$. We would like to find this $\epsilon$-minimizer in the minimum possible wall-clock time. In synchronous federated settings, the total time required to find such a minimizer is the sum of time spent in each round. Each round comprises computation on the server, communication between the server and clients selected for the round, and the local computation on selected clients. For each client $i$, we assume that the total time taken for all computation and communication in each round is its delay, denoted by $\tau_i$. We assume that this is constant across rounds. Further, the time spent on server computation is considered to be negligible in comparison to $\tau_i$. Additionally, we assume that the clients are sorted by delay, so $\tau_i \leq \tau_j, \forall 1 \leq i < j \leq m$ and that all clients are available for selection in all rounds. Using this delay model, the time spent in each round is simply the maximum of the delay of the clients selected in that round. To minimize the time spent in a single round, one would select the clients with the lowest delay.

Our goal, however, is to minimize the total runtime, which is the product of number of rounds with delay per round. Choosing the fastest clients in every round might end up taking more rounds to converge to an $\epsilon$-minimizer thus resulting in a larger total runtime. If we fix the optimization algorithm to the simplest FL algorithm, FedAvg~\cite{pmlr-v54-mcmahan17a}, then the problem of minimizing total runtime becomes a client selection problem. We describe FedAvg with a \texttt{ClientSelection} subroutine in Algorithm~\ref{alg:fedavg}. Selecting the fastest clients every round is a good strategy when all clients have the same data distribution. However, if client distributions are heterogeneous, the solution $w$ might be biased towards the selected clients and thus need more rounds to converge to the $\epsilon$-minimizer of the average loss over all clients (Eq~\eqref{eq:main_opt}). To assess the impact of data heterogeneity in the system theoretically, we use the following assumption.

\begin{assumption}[Pairwise Heterogeneity]\label{assump:pair_het}
For each pair of clients $i,j \in [m]$, there exist constants $B_{i,j},\Gamma_{i,j} \geq 0$  be such that
\begin{align*}
    \norm{\nabla f_i(w) - \nabla f_j(w) }\leq B_{i,j} \norm{\nabla f(w)} + \Gamma_{i,j},\forall w \in \R^d
\end{align*}
\end{assumption}

A similar assumption is widely used~\cite{pmlr-v119-karimireddy20a,pmlr-v130-stich21a} to analyze the convergence of FL algorithms and is referred to as bounded gradient dissimilarity which uses a single $B$ and $\Gamma$ to characterize the heterogeneity of the complete system. We use this pairwise formulation instead, to obtain a finer characterization of heterogeneity between clients. Here, a large $B_{ij}$ or $\Gamma_{ij}$ imply that clients $i$ and $j$ are very different. Instead of pairwise formulation, we could have used a per-client formulation, which puts an upper bounds $\norm{\nabla f_i(w) - \nabla f(w)}$, these two definitions are interchangeable.

As mentioned before, ~\cite{luo_tackling_2022,lat_het,chen2022optimal} utilize a poor characterization of heterogeneity, where different clients have different bounds on the gradient norm, i.e., $\norm{\nabla f_i(w)} 
\leq G_i, \forall i\in [m]$. Using this to bound $\norm{\nabla f_i(w) -\nabla f_j(w)} \leq \norm{\nabla f_i(w)} + \norm{\nabla f_j(w)} \leq G_i + G_j, \forall i,j\in [m]$. Note that this bound is obtained by a trivial and loose triangle inequality which removes any notion of interdependence between clients. Consider the case when clients $i,j$ actually have the same distribution, in which case, our assumption would yield $B_{i,j}=\Gamma_{i,j} =0$. However, using the gradient norm bound would imply, $\Gamma_{i,j} = G_i + G_j= 2G_i\neq 0$ which could be very large indicating that the clients are dissimilar even though they are similar. Further, using only gradient norms does not result in a  non-zero $B_{i,j}$ which as we will see later, is essential to the analysis of total runtime. In Section~\ref{sec:interpretation}, we further clarify how to interpret the above assumption in real loss functions.

From this assumption, we can see that selecting the fastest clients might end up in slower convergence if the fastest clients have high pairwise heterogeneity with respect to the rest of the clients. Further, client selection strategies that accommodate heterogeneity~\cite{balakrishnan2022diverse,pmlr-v151-jee-cho22a} might take fewer rounds to converge but they can select clients with large delay making every round slower. Therefore, to obtain the algorithm with the best runtime, we need to balance both the time taken per round and the number of rounds for convergence. This requires a client selection scheme that balances both the delays and heterogeneity of the clients. In the next section, we derive two such theoretically optimal client sampling schemes.

\textit{Notation :} We use $[m]$ to denote the set $\{1,2,3,\ldots,m\}$ and $S\sim p^K$ to denote a multiset $S$ of size $K$ where each element has been sampled iid from the probability distribution $p=[p_1,p_2,\ldots, p_m]$ on the set $[m]$.

\section{Proposed Client Selection Schemes}
\label{sec:algorithm}
We consider two broad class of client selection schemes where in each round, we either i) select a fixed subset of clients according to some metric, or ii) sample a fixed number of clients according to some distribution.
Note that almost all client selection schemes including ~\cite{balakrishnan2022diverse,reisizadeh_straggler-resilient_2020,pmlr-v151-jee-cho22a,nishio2019client} and even random client selection fall under these two classes. 
We propose two client selection schemes, one for each of the above class of client selection schemes, to cover a broad class of all client selection schemes. We defer all proofs to the Appendix.

To optimize the total runtime of FL algorithms, we need to estimate both the delay per round and the number of rounds to achieve an $\epsilon$-optimizer of $f$. To obtain our proposed client selection schemes, we estimate both these quantities theoretically, to obtain the total theoretical runtime. Our client selection scheme then selects clients to minimize this theoretical runtime. With the following minor modifications, we can estimate these quantities for both these classes of client selection -- for i) by selecting the \textbf{same} subset of clients, and for ii) by sampling from the \textbf{same} probability distribution every round.

The time spent per round is estimated from delay of the clients, while we use the theoretical convergence guarantees of in heterogeneous FL to obtain the number of rounds. For the rest of this section, we focus on the case of no local steps ($E=1$), to simplify our analysis. Further, we deal with a specific class of functions whose convergence can easily be analyzed which are defined in the following assumption.

\begin{assumption}[Loss]\label{assump:loss}
$f$ is $\mu$-P\L{} and $L$-smooth. 
\begin{align*}
 &\norm{\nabla f(w)}^2 \geq\, 2\mu (f(w) - f(w^\star)),\quad \forall w \in \R^d\\
 &\norm{\nabla f(w_1) - \nabla f(w_2)} \leq\, L\norm{w_1 - w_2}, \quad \forall w_1, w_2 \in \R^d
\end{align*}
\end{assumption}
This class of functions is a relaxation of the class of all quadratic functions. Such assumptions are often necessary to derive linear convergence ($\mathcal{O}(\log(1/\epsilon))$ convergence rate) of gradient-based algorithms, like SGD or FedAvg, in centralized and federated settings~\cite{karimi2020linearconvergencegradientproximalgradient,pmlr-v108-bayoumi20a}. The P\L{} condition allows us to handle moderately non-convex losses. Further, all $\mu$-strongly convex losses, which includes quadratics, satisfy the P\L{} condition. Note that P\L{} functions are generally non-convex functions, and include certain deep neural network architectures~\cite{LIU202285}.

\subsection{\texttt{DelayHetSubmodular}}
\label{sec:submodular}
\begin{algorithm}[t!]
\begin{algorithmic}
\FORALL{clients $i\in [m]$}
\STATE $A_i \gets$ estimated feature covariance from client $i$.
\STATE $\tau_i \gets$ estimated delay from client $i$
\ENDFOR
\STATE $A  \gets \frac{1}{m}\sum_{i\in m} A_i$
\STATE $B_{i,j} \gets \norm{(A_i - A_j)A^{-1}},\, \forall i,j\in [m]$
\STATE $S \gets \min_{S \subseteq [m]} \frac{\max_{i\in S}\tau_i}{1 - 2(\frac{1}{m}\sum_{j=1}^m (\min_{i\in S} B_{i,j}))^2}$
\STATE $\alpha_i \gets \frac{1}{m}\sum_{j=1}^m \I{i = \arg\min_{i'\in S} B_{i',j}}$
\STATE Return $S, \{\alpha_i\}_{i\in S}$
\end{algorithmic}
    \caption{\texttt{DelayHetSubmodular}}
\label{alg:submodular}
\end{algorithm}

For our first scheme, we assume that the selected set of clients and coefficients from the \texttt{ClientSelection} subroutine in Algorithm~\ref{alg:fedavg} is constant across rounds. 
Let this set be $S$. We will derive the theoretical number of rounds and the theoretical delay in each round for this set. Note that each round takes $\max_{i\in S} \tau_i$ time to run. This provides us with the delay.

To analyze the theoretical number of rounds, we need to obtain convergence in the global loss function $f$ (Eq~\eqref{eq:main_opt}), while only using the gradients from clients in set $S$. We will show that by appropriately choosing the coefficients $\{\alpha_i\}_{i\in S}$, the gradient of clients in the selected set $S$, denoted as $\nabla f_S(w) \triangleq \sum_{i\in S} \alpha_i \nabla f_i(w)$ is a good approximation to the true gradient $\nabla f(w)$. Therefore, doing gradient descent with $\nabla f_S(w)$ should approximately minimize $f(w)$ providing us with an approximate $\epsilon$-minimizer and the number of rounds to reach it. To establish that $\nabla f_S(w)$ is a good approximation to $\nabla f(w)$, we need to first choose the coefficients $\{\alpha_i\}_{i\in S}$ to define $\nabla f_S$ (Definition~\ref{def:coeff}), then we need to compute the bias of $\nabla f_S$ compared to $\nabla f$(Lemma~\ref{lem:bias_grad_submod}) and finally, we need conditions on when this bias is small(Assumption~\ref{assump:bdd_het}) to allow enough for theoretical convergence and computing the number of rounds(Theorem~\ref{thm:submodular}).

The first task in obtaining a good approximation of the true gradient $\nabla f(w)$ by $\nabla f_S(w)$ is choosing coefficients $\{\alpha_i\}_{i\in S}$. The intuition behind selecting coefficients is simple. Consider the difference $\norm{\frac{1}{m}\sum_{i=1}^m \nabla f_i(w) - \sum_{j\in S}\alpha_j f_j(w)}$. To minimize this difference, for each client $j\in [m]$ not present in the set $j\notin S$, we use one of the selected clients $i\in S$ as a proxy. A good proxy for client $j$ would be a client $i\in S$ such that $\arg\min_{i\in S}\norm{\nabla f_i(w) - \nabla f_j(w)}$. From our heterogeneity assumption ~\ref{assump:pair_het}, this difference is bounded by $B_{i,j}\norm{\nabla f(w)} + \Gamma_{i,j}$, thus we should select $i\in S$ which minimizes this upper bound. Instead of the complete upper bound, we choose $i$ based on the term $B_{i,j}$, which is the only term affecting the speed of convergence. Therefore, the proxy of client $j$ among selected clients is denoted by $\beta_j(S)$. The coefficient for any selected client is simply the number of clients of which it is a proxy divided by $m$. The following definition formally states this choice of coefficients.

\begin{definition}[Coefficients]\label{def:coeff}
For any set $S\subset [m]$, the coefficients $\{\alpha_i\}_{i\in S}$ are    $\alpha_i = \frac{1}{m}\sum_{j=1}^m \I{\beta_j(S) = i}\text{, where } \beta_j(S) = \argmin_{i\in S} B_{i,j}$.
\end{definition}
Choosing a proxy for each client that has not been selected is not a new idea, and has been used earlier for coreset selection~\cite{pmlr-v119-mirzasoleiman20a} and in DivFL~\cite{balakrishnan2022diverse}. However, DivFL considers the full gradient, while we only consider the term $B_{i,j}$ which affects speed of convergence.

With this choice of coefficients, we can quantify the bias of using $\nabla f_S(w)$ instead of the full gradient $\nabla f(w)$. If this bias, denoted by $\norm{\nabla f_S(w) - \nabla f(w)}$ is low, we can utilize existing results on convergence with biased gradients~\cite{biased_grad_framework}, to obtain number of rounds using $\nabla f_S(w)$. The following lemma establishes this bias in terms of heterogeneity terms $B_{i,j}$'s and $\Gamma_{i,j}$'s.

\begin{lemma}[Biased Gradient]\label{lem:bias_grad_submod}
For any set $S$, with coefficients defined by Definition~\ref{def:coeff}, $\nabla f_S$ is a biased gradient of $S$, such that $\norm{\nabla f_S(w) - \nabla f(w)}^2 \leq \,  B_S \norm{\nabla f(S)}^2 + \Gamma_S, \,\, \forall w \in \R^d$,where  $B_S = 2(\frac{1}{m}\sum_{j=1}^m B_{j, \beta_j(S)})^2,  \Gamma_S = 2(\frac{1}{m}\sum_{j=1}^m \Gamma_{j, \beta_j(S)})^2$.
\end{lemma}
The above lemma is a simple application of the triangle inequality and Assumption~\ref{assump:pair_het}. It shows that the bias term is actually bounded by terms of $B_S$ and $\Gamma_S$ resembling those Assumption~\ref{assump:pair_het}, however, now decided by the choice of the selected set $S$. We can already see that the bias is low, whenever $B_S$ and $\Gamma_S$ is small.

To establish convergence with $\nabla f_S(w)$, we need the bias defined above to be small. Specifically, as long as $B_S < 1$, $\nabla f_S(w)$ is still a good approximation to $\nabla f(w)$ , which necessitates the following assumption bounding the data heterogeneity that our methods can tolerate.

\begin{assumption}(Bounded Heterogeneity)\label{assump:bdd_het}
    Let $\max_{i\in [m]}\frac{1}{m}\sum_{j=1}^m B_{i,j} < 1/ \sqrt{2}$.
\end{assumption}
To prove that this assumption implies small bias, i.e. $B_S <1$, note that $B_S$ is maximized when $S$ is a singleton as $B_S$ is a decreasing function of $S$ for $\abs{S}>0$. The above assumption ensures that whenever $S$ is a singleton, it's bias is $< 1$, and in turn $B_S <1, \forall S\subseteq [m]$.

Now, we have all the required assumptions to establish convergence using $\nabla f_S(w)$ to an $\epsilon$-minimizer. We use existing convergence results for the biased gradient framework  \citep[Theorem~6]{biased_grad_framework} to obtain the number of rounds.

\begin{theorem}\label{thm:submodular}
    Under Assumptions~\ref{assump:pair_het},~\ref{assump:loss} and ~\ref{assump:bdd_het}, running Algorithm~\ref{alg:fedavg} for $R$ rounds using a constant stepsize $\eta \leq\frac{1}{L}$, with a fixed client set $S_r=S \subseteq [m], \forall r \in [R]$, coefficients $\{\alpha_i\}_{i\in S}$ defined in Definition~\ref{def:coeff}, for
       $ R  \geq \frac{L}{\mu(1 - B_S)}\log\bigg(\frac{f(w_0) - f(w^\star)}{\epsilon}\bigg)$,
   we obtain $f(w_R) - f(w^\star) \leq \epsilon + \frac{\Gamma_S}{\mu(1-B_S)}$.
\end{theorem}
From the above theorem, by using a fixed subset $S$ of clients, we cannot converge to an $\epsilon$-minimizer of $f$, rather only to its neighborhood. The size of this neighborhood is determined by $\Gamma_S$, which depends on $\Gamma_{ij}$. The number of rounds required to converge to this neighborhood, however, depends only on the term $B_S$ and thus only $B_{i,j}$'s. Therefore, if we select a set of clients having small data heterogeneity leading to small $B_S$, we will require fewer number of rounds to converge. Note that $B_S$ is small, when the selected clients is a diverse and representative subset of all the clients. Further, as only $B_S$ appears in convergence rate, our choice of coefficients in Definition~\ref{def:coeff} based on only $B_{ij}$ is justified.

Our analysis provides new insight into the commonly used heterogeneity assumption~\ref{assump:pair_het}, -- while $\Gamma_{ij}$ decides the final convergence error, $B_{ij}$ decides the slowdown in convergence rate. This connection can only be established by using biased gradient framework (Lemmas~\ref{lem:bias_grad_submod}).

Now that we have the theoretical number of rounds and the theoretical delay per round, we can write down the theoretical total runtime for a given set of clients $S$.
\begin{remark}[Runtime]
    The total runtime to achieve an $(\epsilon + \frac{\Gamma_S}{\mu(1-B_S)})$-minimizer of $f$ for any subset $S\subseteq[m]$ is $\mathcal{O}(\frac{\max_{i\in S} \tau_i L }{\mu(1 - B_S)}\log(1/\epsilon))$. To minimize the total runtime, we need to find the subset $\min_{S \subseteq[m]} g(S) \triangleq \frac{\max_{i\in S} \tau_i}{(1 - B_S)}$
\end{remark}
The total runtime is simply the product of number of rounds and constant delay per round. The function $g(S)$ precisely encodes the relationship between the data heterogeneity and latency heterogeneity to total runtime. Therefore, the set of clients $S$ that minimize it are theoretically optimal in terms of total runtime, and should outperform simple heuristic-based baselines. Minimizing $g(S)$ is the core-idea behind our first algorithm, \texttt{DelayHetSubmodular}, described in Algorithm~\ref{alg:submodular}.

Now, that we know minimizing $g(S)$ gives us theoretical optimality, we need to see if this is actually computationally possible. Note that $S\subseteq [m]$, therefore there are $2^m$ possible choices for $S$. A naive approach for minimizing $g(S)$ is to search over all possible $S$, which would require $\mathcal{O}(2^m)$ time making our algorithm computationally inefficient. To alleviate this, we utilize the intricate structure of the objective $g$ by the following remark.

\begin{remark}[Submodular]\label{rem:submodular}
    $g(S)$ is submodular in $S$.
\end{remark}
Submodular functions, formally defined in ~\cite{submodmin}, are a large class of functions on subsets whose properties closely convexity. As we want to minimize a submodular function, we can use existing  efficient minimization algorithms~\cite{submodmin,submodmin_polytime,submodmin_proof}, that require $\mathcal{O}(\mathrm{poly}(m))$, which is much better than the naive $\mathcal{O}(2^m)$ time.

Note that no existing baseline apart from ~\cite{luo_tackling_2022,lat_het} minimizes total runtime, focusing on sampling variance~\cite{chen2022optimal} or convergence error~\cite{balakrishnan2022diverse,pmlr-v151-jee-cho22a}. Further, none of these works consider the distinction between terms affecting convergence error and convergence speed. Therefore, theoretically, our method provides a clearer picture of convergence under data and latency heterogeneities than existing works.

\subsection{\texttt{DelayHetSampling}}
\label{sec:sampling}
 For our second scheme, the \texttt{ClientSelection} subroutine, we fix a probability distribution over clients   $p$. Following the methodology from previous section, we compute the theoretical delay per round and the number of rounds theoretically under this sampling distribution. This approach is closest to ~\cite{luo_tackling_2022,lat_het}, with the key difference being the modeling of heterogeneity assumptions.

Formally, we sample a multiset of $K$ clients with replacement from a probability distribution $p = [p_1,p_2,\ldots, p_m]$ over the set of clients, where $\sum_{i=1}^m p_i = 1$ and $p_i \in [0,1],\forall i\in [m]$. Let $S_r \sim p^K$ be the multiset of clients sampled at round $r$. We set the coefficients to be $\alpha_i = \frac{1}{K}, \forall i \in S_r$. As we are sampling a multiset, the same client might be sampled more than once. In that case, each time that it is sampled, it obtains a coefficient $\frac{1}{K}$ and we sum them up to get its overall coefficient. 

\begin{algorithm}[t!]
\begin{algorithmic}
\FORALL{clients $i\in [m]$}
\STATE $A_i \gets$ estimated feature covariance from client $i$
\STATE $\tau_i \gets$ estimated delay from client $i$
\ENDFOR
\STATE $A  \gets \frac{1}{m}\sum_{i\in m} A_i$
\STATE $B_{i,j} \gets \norm{(A_i - A_j)A^{-1}}, \forall i,j\in [m]$.
\STATE Let $\tilde{B}\in \R^{m\times m}$ where $(\tilde{B})_{ij} = B_{ij}^2$.
\STATE $B_p \gets 2(\frac{p^\intercal \tilde{B}\mathbf{1}_m}{m} + \frac{p^\intercal \tilde{B}p}{K})$
\STATE $p^\star \gets \min_{p \in [0,1]^m, \sum_{i}p_i=1}\frac{\sum_{i=1}^m ((\sum_{j=1}^{i}p_j)^K - (\sum_{j=1}^{i}p_j)^K)\tau_i}{1 - B_p}$ 
\STATE $S\gets$ Sample a multiset of $K$ clients from the probability distribution $\mathbf{p}^\star$ with replacement.
\STATE $\alpha_i = \frac{1}{K}, \quad \forall i \in S$
\STATE Return $S,\{\alpha_i\}_{i\in S}$
\end{algorithmic}
    \caption{\texttt{DelayHetSampling}}
\label{alg:optimalsampling}
\end{algorithm}

 Since we are dealing with a sampling scheme, we compute the expected number of rounds and the expected delay under the sampling distribution $p$.
 \begin{lemma}[Expected Delay]\label{lem:exp_delay}
     The expected delay of sampling $K$ clients from $p$ is given by
     $\E_{S \sim p^K} [\max_{i\in S} \tau_i] = \sum_{i=1}^m [(\sum_{j=1}^{i}p_j)^K-(\sum_{j=1}^{i-1} p_j)^K]\tau_i$
 \end{lemma} 
Since $\tau_i$ are in ascending order, probability that the maximum delay is $\leq \tau_i$ is $(\sum_{j=1}^{i} p_j)^K$. The probability that it is exactly equal to $\tau_i$ is obtained by subtracting the probability that it is $\leq \tau_{i-1}$.

Computing theoretical number of rounds to reach an $\epsilon$-minimizer with sampling distribution $p$, we follow a similar strategy to previous section. In each round, we use the gradient $\nabla f_{S_r}(w) = \frac{1}{K}\sum_{i \in S_r} \nabla f_i(w)$. We will show that this is an approximation $\nabla f(w)$, and quantify  the error in this approximation in Lemma~\ref{lem:expected_bias}. Then, we establish conditions under which this approximation is good enough(Assumption~\ref{assump:bdd_het_sampl}) so that we can use biased gradient framework as before~\cite{biased_grad_framework} to obtain number of rounds theoretically in Theorem~\ref{thm:sampling}.

  The following lemma bounds the expected bias of the sampling gradient wrt the original gradient $\nabla f(w)$.
\begin{lemma}(Expected Bias)\label{lem:expected_bias}
The expected bias of a multiset $S \sim p^K $ of $K$ clients with coefficients $\alpha_i=\frac{1}{K}, \forall i \in S$ and $\forall w \in \R^d$ is $\E_{S\sim p^K}[\norm{\nabla f_S(w) - \nabla f(w)}^2]\leq B_{p}\norm{\nabla f(w)}^2 + \Gamma_p$, where  $B_p = 2(\frac{p^\intercal \tilde{B} \mathbf{1}_m}{m}+ \frac{p^\intercal \tilde{B}p}{K}), \, \Gamma_p = 2(\frac{p^\intercal \tilde{\Gamma} \mathbf{1}_m}{m}+ \frac{p^\intercal \tilde{\Gamma}p}{K})$ and  $\tilde{B}, \tilde{\Gamma} \in \R^{m\times m}$ are matrices with entries $\tilde{B}_{ij} = (B_{ij})^2$ and $\tilde{\Gamma}_{ij} = (\Gamma_{ij})^2$ and $\mathbf{1}_m \in \R^m$ has all its entries as $1$. 
\end{lemma}
As we observed in Lemma~\ref{lem:bias_grad_submod}, the bias here also depends on $B_p$ and $\Gamma_p$ resembling those in Assumption~\ref{assump:pair_het}, with the exact dependence decided by the sampling distribution $p$. Again, the bias is low whenever $B_p$ and $\Gamma_p$ are small. For the sample gradient to be a good approximation to $\nabla f(w)$, we require $B_p<1$, which allows us to apply the biased gradient framework~\cite{biased_grad_framework}. The following bound on data heterogeneity ensures this.

\begin{assumption}\label{assump:bdd_het_sampl}
    Let $\max_{i\in [m]}\frac{1}{m}\sum_{j=1}^m B_{ij}^2  < \frac{1}{2}$.
\end{assumption}
To prove that this assumption implies $B_p<1$, we find the maximum value of $B_p$ for any distribution $p$. As $B_p$ is a quadratic in $p$, its maxima over $p$ is achieved at one of the corners of the probability simplex, i.e., $p_i=1$ for some $i\in [m]$. The above assumption ensures that at any of these corners, $B_p<1$, implying that $B_p<1, \forall p$.

Note that Assumption~\ref{assump:bdd_het} implies Assumption~\ref{assump:bdd_het_sampl} by Jensen's inequality, thus our client sampling strategy can handle more data heterogeneity than submodular client selection.

We now have all the required assumptions to apply biased gradient framework, and we adapt the results of  ~\citep[Theorem~6]{biased_grad_framework} to obtain the theoretical number of rounds for the sampling scheme $p$.
\begin{theorem}\label{thm:sampling}
    Under Assumptions~\ref{assump:pair_het},\ref{assump:loss} and ~\ref{assump:bdd_het_sampl}, running Algorithm~\ref{alg:fedavg} for $R$ rounds with constant stepsize $\eta\leq \frac{1}{L}$, where a multiset of clients $S_r\sim p^K$ of $K$ clients with coefficients $\alpha_i = \frac{1}{K}, \forall i\in S_r$ is sampled every round $r \in [R]$, if  $R  \geq \frac{L}{\mu(1 - B_p)}\log\bigg(\frac{f(w_0) - f(w^\star)}{\epsilon}\bigg)$,
    we obtain $\E[f(w_R)] - f(w^\star) \leq \epsilon + \frac{\Gamma_p}{\mu(1-B_p)}$.
\end{theorem}
Similar to Theorem~\ref{thm:submodular}, convergence in expectation is only possible to a neighborhood of the optima determined by $\Gamma_p$, while the number of rounds required is decided by $B_p$. Therefore, a sampling distribution $p$ obtaining small $B_p$ should provide smaller number of rounds.

Using this, we can now compute the total runtime theoretically.

\begin{remark}[Runtime]\label{rem:sampling}
    The total expected runtime to achieve an $(\epsilon + \frac{\Gamma_p}{\mu(1-B_p)})$-minimizer on expectation $f$ for probability distribution $p$ is $\mathcal{O}(\frac{ \sum_{i=1}^m [(\sum_{j=1}^{i}p_j)^K-(\sum_{j=1}^{i-1} p_j)^K]\tau_i L }{\mu(1 - B_p)}\log(1/\epsilon))$. To minimize the total runtime, we need to find the sampling distribution $\min_{p} g(p) \triangleq \frac{\sum_{i=1}^m [(\sum_{j=1}^{i}p_j)^K-(\sum_{j=1}^{i-1} p_j)^K]\tau_i L }{\mu(1 - B_p)}$
\end{remark}

Note that this minimization is over a constraint set, the probability simplex $\Delta_m = \{p_i \geq 0, \forall i\in [m], \sum_{i=1}^m p_i=1\}$. Although, the objective function $g(p)$ is not convex in $p$, the constrained set is simple, and we can use interior point methods~\cite{trust_constr} for minimizing it. This forms the core-idea of our second client selection scheme described in Algorithm~\ref{alg:optimalsampling}.

\subsection{Computational Efficiency}
In this section, we discuss the related computational requirements of our methods and closest baselines ~\cite{luo_tackling_2022,lat_het}. Note that using submodular minimization and interior point methods for Algorithms~\ref{alg:submodular} and ~\ref{alg:optimalsampling} respectively yields computationally efficient methods. 
Compared to this ~\cite{luo_tackling_2022,lat_het} can also utilize interior point methods to minimize their estimates of theoretical runtime similar to our sampling scheme.

However, one of the key issues in minimizing the theoretical runtimes in these papers (~\citep[Algorithm~2]{luo_tackling_2022} and ~\citep[Eq~(P3)]{lat_het}) is that their theoretical runtimes require estimates of non-trivial quantities depending on the loss or model weight suboptimality, like $\frac{\alpha}{\beta}$ in ~\citep[Eq~(22)]{luo_tackling_2022} and $\alpha$ in ~\citep[Eq~(25)]{lat_het}. To estimate loss or model suboptimality, these methods need to estimate an approximation for the optimal loss value or optimal model weight. Note that the goal of FL is to actually obtain optimal model weights or loss (Eq~\eqref{eq:main_opt}). So, these papers need to partially solve the full FL problem, via warm-up steps, before even attempting to solve a client selection subroutine. These warm-up steps require full client participation adding an expensive delay overhead, and even then obtain rough approximations of the suboptimality as obtaining good approximations of suboptimality would imply actually minimizing the FL objective (Eq~\eqref{eq:main_opt}). 

A natural question here would be whether our methods require similar suboptimality estimation. Note that this is not the case as we only require estimates of $B_{i,j}$ and $\Gamma_{i,j}$ to run our methods which depend on norm of gradient difference. This can be calculated using only $1$ warm-up round. To further improve the estimation of $B_{i,j}$ and $\Gamma_{i,j}$, we discuss an important connection of these terms to different forms of heterogeneity in the next section.

\section{Interpreting Data Heterogeneity via Linear Regression}
\label{sec:interpretation}

In this section, we provide an interpretation of the terms $B_{i,j}$ and $\Gamma_{i,j}$ used in Assumption~\ref{assump:pair_het}. 
When implementing our algorithms in practice, it is easy to estimate the norm of gradient difference $\norm{\nabla f_i(w) - \nabla f_j(w)}$, however, this would be an estimate of $B_{i,j}\norm{\nabla f(w)} + \Gamma_{i,j}$. It is unclear how to obtain tight estimates of $B_{i,j}$ and $\Gamma_{i,j}$ from this sum as we only have the bounded heterogeneity requirements (Assumptions~\ref{assump:bdd_het} and ~\ref{assump:bdd_het_sampl}) for maximum tolerable $B_{i,j}$'s. Further, for deep neural networks, the gradients are high dimensional, so the difference of gradients in these cases is then the difference of two high dimensional vectors. Difference of two high dimensional vectors are roughly constant, making tight estimation of $B_{i,j}$, an integral component of our algorithms,  difficult.

To alleviate these difficulties, we utilize the example of of linear regression, introduced in Section~\ref{sec:motivation}, to interpret the terms $B_{ij}$ and $\Gamma_{ij}$. Note that the linear regression interpretation is still valid for high dimensional neural networks, as in the NTK regime~\cite{ntk_jacot,chizat_bach}, neural networks are roughly linear in their features.

We now formally define the linear regression setup.
If $\{(x_{i,q}, y_{i,q})\}_{q\in [n]}$ are the $n$ datapoints on a client $i\in [m]$, without any label noise, then using $l(w, (x,y) = \frac{1}{2}(y - \ip{w}{x})^2$ as the squared loss per datapoint, we can compute the training loss $f_i$ for that client.    $f_i(w) = \frac{1}{n}\sum_{j=1}^n l(w, (x_{i,j}, y_{i,j})) = \frac{1}{2} (w - w_i^\star)^\intercal A_i (w -w_i^\star)$
where $A_i=\frac{1}{n}\sum_{q=1}^n x_{i,q} x_{i,q}^\intercal$ is the empirical covariance matrix of the features client $i$. Then, $f(w) = \frac{1}{2} (w - w^\star)^\intercal A (w - w^\star) + \frac{1}{m}\sum_{i\in [m]}f_i(w^\star)$.
 Here,  $A = \frac{1}{m}\sum_{i\in [m]} A_i$, and $w^\star = \frac{1}{m}\sum_{i\in [m]}A_i w_i^\star$ are the average feature covariance of all clients and the minimizer of $f$ respectively. By plugging in these expressions into Assumption~\ref{assump:pair_het}, we obtain the values of $B_{ij}$ and $\Gamma_{ij}$.

\begin{proposition}[Pairwise Heterogeneity  Linear Regression]\label{lemma:pair_het_lin_reg}
The linear regression model satisfies Assumption~\ref{assump:pair_het} with  
\begin{align*}
     B_{i,j} = \norm{ (A_i - A_j)A^{-1}}_2\quad \text{   and   }\quad\Gamma_{i,j} = \norm{A_i(w^\star - w_i^\star)  - A_j (w^\star -  w_j^\star)}_2, \quad\quad\forall i,j \in [m].
\end{align*}  
\end{proposition}
Here, $\norm{A}_2 = \sigma_{\max}(A)$ is the maximum singular value, for a $d \times d$ matrix.  Since all $A_i, i\in [m]$ and $A$ are symmetric PSD matrices, $\norm{A}_2 = \sigma_{\max}(A)=\lambda_{\max}(A)$. 
With this lemma, we can interpret the two sources of heterogeneity in Assumption~\ref{assump:pair_het}
-- 
\begin{enumerate}
    \item  \textbf{Feature Heterogeneity ($B_{ij}$)} : This corresponds to different clients having different feature covariances and this affects the speed of convergence, 
    \item \textbf{Optima Heterogeneity ($\Gamma_{ij}$)} :  This depends on both the client minimizers $w_i^\star$ and their feature covariances and is $0$ and this affects the final convergence error. When all clients have the same optima, $w_i^\star = w^\star, \forall i\in [m]$, this term also depends on $\norm{A_i - A_j}_2$.
\end{enumerate}

With this interpretation, estimating $B_{ij}$ for Algorithms~\ref{alg:submodular} and \ref{alg:optimalsampling} boils down to  estimating the feature heterogeneity for each client. Further, estimating Feature heterogeneity is possible even for non-linear models, like kernels or deep neural networks, using an appropriate definition of features, for instance the last layer activations in deep neural networks. Note that this interpretation separates our algorithm from ~\cite{chen2022optimal, balakrishnan2022diverse,luo_tackling_2022}, which use only the empirical gradient for estimation, and thus suffer from the issues in high dimensions. We now validate the performance of our schemes via experiments.

\section{Experiments}
\label{sec:add_experiments}
In this section, we compare our proposed client selection methods against baselines for FedAvg.

\subsection{Setup}
Here, we provide a detailed description of our experimental setup.

\begin{table*}[t!]
    \small
    \centering
        \resizebox{\textwidth}{!}{
    \begin{tabular}{cc|ccccccc}
    \toprule
    Datasets & Target  & \makecell{\texttt{Submodular}\\
    Alg.~\ref{alg:submodular}}
    & \makecell{\texttt{Sampling}\\
    Alg.~\ref{alg:optimalsampling}} & Random & \makecell{FLANP\\~\cite{reisizadeh_straggler-resilient_2020}} & \makecell{DivFL\\~\cite{balakrishnan2022diverse}} & \makecell{Power-of-\\Choice~\cite{pmlr-v151-jee-cho22a}}&\makecell{ApproxOpt-\\Sampling\\~\cite{lat_het}}\\
    \midrule
    Quadratic &$2.95$ & \textbf{0.57}	& 0.593 &	0.714	&3.5	&0.635	& 0.68 &0.803 \\
    Dirichlet MNIST& $0.95$ & 0.687	& 0.224 &	0.586& 	\textbf{0.162}&	0.491 &	0.5& 0.582\\
    Rotated MNIST &$0.93$ &0.845	&0.513&	1.122	&\textbf{0.36}&	1.186 &	0.804 & 0.605    \\
    Dirichlet FashionMNIST &$0.83$ & 1.672	&	\textbf{0.966}	&2.773	&1.694&	2.597	&2.961 & 4.038\\
    Rotated FashionMNIST& $0.77$ &1.635	&	\textbf{0.795}	&4.142&	1.097	&3.304&	1.876& 3.432\\
    Dirichlet CIFAR-10 &$0.67$ &\textbf{1.1}&	1.516	&1.102	&1.188	&	1.12	&1.106 &1.557\\
    Rotated CIFAR-10 & $0.66$ & \textbf{1.436}	&	2.43&	2.296& N/A		&2.928	&1.896&1.794\\
    FEMNIST & $0.8$ & 14.68	&	\textbf{4.75}	&24.98&	19.75 &	29.045 &	64.2& N/A\\
    Shakespeare & $0.43$ & 0.8353	&	\textbf{0.807}&	3.055&	3.682&	3.549	&1.857&1.217\\
    \bottomrule\\
    \end{tabular}
    }
    \caption{\normalsize Total runtime (in $10^3$ seconds) to reach target test accuracy/loss with synthetic delay distribution. For Quadratic dataset, we use test normalized loss, for other datasets, we use test accuracy. The lowest runtime for each dataset is \textbf{bold}. N/A denotes that the algorithm did not converge to target accuracy/loss.}    
    \label{tab:uniform_runtime}
    
\end{table*}

\paragraph{Datasets}

\textbf{Quadratic}: We implement the linear regression example introduced in Section ~\ref{sec:motivation}.  Feature heterogeneity is obtained by sampling the eigenvalues of the feature covariance matrix from the range $[1,10]$ for each client. Each coordinate of the optimal model $w^\star$ is generated from a $Bernoulli(0.5)$ distribution. The noise in labels has a standard deviation of $0.001$. Additional details on the setup can be found in Section~\ref{sec:motivation}.

\textbf{Simulated Datasets}: We split MNIST~\cite{deng2012mnist}, FashionMNIST~\cite{xiao2017fashionmnist} and CIFAR10~\cite{cifar10} into $m=100,100,50$ clients respectively. To add data heterogeneity, we use two strategies-- i) Dirichlet and ii) Rotated. In Dirichlet heterogeneity, we sample the fraction of labels on each client from a $Dirichlet(\alpha)$ distribution with $\alpha=2$, where $\alpha$ controls the label distribution. For $\alpha=\infty$, each client has uniform label distribution, while for $\alpha=1$, the label distribution on each client is sampled uniformly from a simplex. In Rotated heterogeneity, there is no label imbalance, but, the images on each client are rotated by an angle sampled uniformly from $[0^\circ,45^\circ]$. This ensures feature heterogeneity. We use $2$-layer ConvNet for MNIST and FashionMNIST and ResNet18~\cite{resnet} for CIFAR10.

\textbf{Real Datasets}: We use two real federated datasets, FEMNIST and Shakespeare from the LEAF benchmark~\cite{leaf}. In FEMNIST, each client contains images drawn by a particular individual. Shakespeare is dataset for next character prediction, which involves dialogues from Shakespeare's plays. Each client contains dialogues of a particular character. We use a $2$-layer ConvNet for FEMNIST and a StackedLSTM for Shakespeare.
\paragraph{Implemenentation details for Algorithms~\ref{alg:submodular} and \ref{alg:optimalsampling}}
For each dataset, we use a single batch to compute empirical feature covariance. For Quadratic dataset, we use the raw features. For image and text datasets, raw features are very high-dimensional, so, we use the input to the last layer of their models as features. This insight has been used in transfer learning~\cite{transfer_learning}. We ensure that $B_{i,j}$ values satisfy Assumption~\ref{assump:bdd_het}. $B_{i,j}$ are updated every round for the clients which participated in that round, akin to DivFL~\cite{balakrishnan2022diverse}.
For submodular minimization we use an implementation of MinNorm algorithm~\cite{submodmin}. The MinNorm algorithm requires the value of the submodular function for the empty set. From Remark~\ref{rem:submodular}, this value is $\infty$. For numerical stability, we set it to the maximum value of $g(S)$, when $S$ is a singleton. For sampling, we solve an approximation to $g(p)$ by setting $K=1$. Similar approximations have been used previously in ~\cite{luo_tackling_2022} to simplify the minimization procedure.
Further, we use scipy~\cite{2020SciPy-NMeth} with 'trust-constr'~\cite{trust_constr} solver for minimizing $g(p)$. Note that we run Algorithms~\ref{alg:submodular} and ~\ref{alg:optimalsampling} every round instead of only once at the start as derived for theory.

\paragraph{Baselines}
We compare our algorithm to $4$ baselines -- random, DivFL~\cite{balakrishnan2022diverse}, Power-of-Choice~\cite{pmlr-v151-jee-cho22a} and FLANP~\cite{reisizadeh_straggler-resilient_2020}. The random baseline samples clients uniformly. DivFL and Power-of-Choice are client selection schemes to tackle data heterogeneity, while FLANP tackles client latency heterogeneity. For all baselines, we use FedAvg with $5$ local epochs as the FL algorithm. We do not compare with ~\cite{luo_tackling_2022}, rather with it's follow-up ~\cite{lat_het}, which we refer to as ApproxOptSampling. For ApproxOptSampling, we compute the sampling distribution at every round similar to our methods. Further, instead of running expensive warm-up steps for ApproxOptSampling, we run a random search to tune it's hyperparameters. Additionally,  to obtain the sampling distribution in ApproxOptSampling, we utilize the same method as our sampling baseline, namely scipy with 'trust-constr' solver. We fix the number of iterations in this solver, so that computing the probability distribution in each round takes at most $1$ hour.

\paragraph{Delay distributions}
We evaluate all methods under two typical delay distributions: synthetic delays and NYCMesh-simulated delays. 
For synthetic delays, we separately sample the communication and computational latency, both in a way that mimics the real-world scenarios.
First, we stochastically sample the link speed, denoted as $c_i$, for each client $i$, from a uniform distribution ranging between 200 KB/s and 5 MB/s. The communication latency is then estimated as $\tau_{i,comm} = MS / c_i$, where $MS$ represents the model size. The computational delay follows a uniform distribution, $\tau_{i,comp} \sim U(15, 100)$. As a result, the latency for selecting client $i$ in a global round is determined by $\tau_i = \tau_{i,comm} + \tau_{i,comp}$.
The procedure for simulating delays in NYCMesh is elaborated in Section~\ref{sec:motivation}.

\begin{table*}[t!]
    \small
    \centering
    \resizebox{\textwidth}{!}{
    \begin{tabular}{cc|ccccccc}
        \toprule
    Datasets & Target  & \makecell{\texttt{Submodular}\\
    Alg.~\ref{alg:submodular}}
    & \makecell{\texttt{Sampling}\\
    Alg.~\ref{alg:optimalsampling}} & Random & \makecell{FLANP\\~\cite{reisizadeh_straggler-resilient_2020}} & \makecell{DivFL\\~\cite{balakrishnan2022diverse}} & \makecell{Power-of-\\Choice~\cite{pmlr-v151-jee-cho22a}}&\makecell{ApproxOpt-\\Sampling\\~\cite{lat_het}}\\
    \midrule 
    Quadratic &$2.95$ & \textbf{1.67} &	7.109 &	63.27	&76.562	& 49.08 &	62.72 &39.96\\
    Dirichlet MNIST & $0.95$ & 1.596	& \textbf{0.992}&	70.35	&1.022 &	31.91&	34.23 & 23.37\\
    Rotated MNIST &$0.93$ & \textbf{2.499}	& 5.356	&114.3	&2.694	& 102.1 &50.60 &46.34\\
    Dirichlet FashionMNIST &$0.83$ & 18.81 & 8.63	&190.3	& \textbf{5.838} &	113.6 &	155.6& 123.6\\
    Rotated FashionMNIST& $0.77$ & 27.86 &	\textbf{14.74} &252.3 &	22.1	& 179.5&130.1& 89.51\\
    Dirichlet CIFAR-10 &$0.67$ & 4.635 &	\textbf{3.297}	& 32.94	& N/A & 43.36 &	34.21 & 25.55\\
    Rotated CIFAR-10 & $0.66$ & \textbf{4.328} &	4.944 &56.57 & N/A &	77.51 &51.56&48.86\\
    FEMNIST & $0.8$ &\textbf{52.25} & 52.52	&1029	&	79.363&	307.5&	1125 & N/A\\
    Shakespeare & $0.43$ &\textbf{3.11}	&6.965&	127.2	&467.2	&147 & 85.53& 73.03 \\
    \bottomrule \\
    \end{tabular}
    }
    \caption{\normalsize Total runtime (in $10^3$ seconds) to reach target test accuracy/loss with NYCMesh delays. For Quadratic dataset, we use test normalized loss, for other datasets, we use test accuracy. The lowest runtime for each dataset is \textbf{bold}. N/A denotes that the algorithm did not converge to target accuracy/loss.}
    \label{tab:nycmesh_runtime}
\end{table*}

\begin{figure*}[t!]
\centering
\normalsize
\subfloat[\normalsize FEMNIST, synthetic delays]{\includegraphics[width=0.45\textwidth]{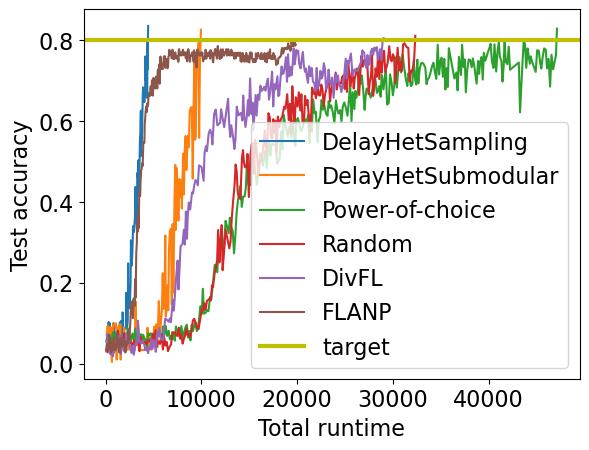}
\label{fig:femnist_uniform}}
\hfill
\subfloat[\normalsize Shakespeare, synthetic delays]{\includegraphics[width=0.45\textwidth]{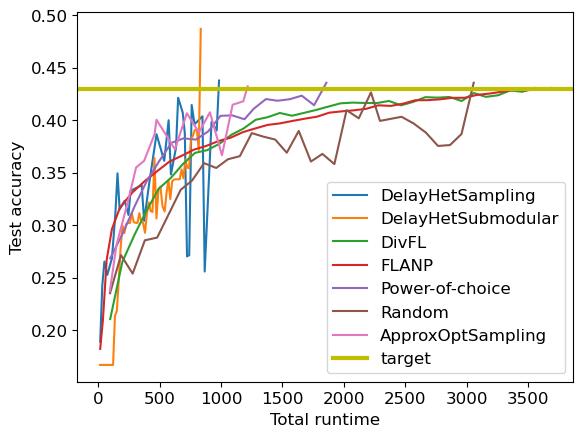}
\label{fig:shakespeare_uniform}}
\vfill
\subfloat[\normalsize Dirichlet FashionMNIST, NYCMesh delays]{\includegraphics[width=0.45\textwidth]{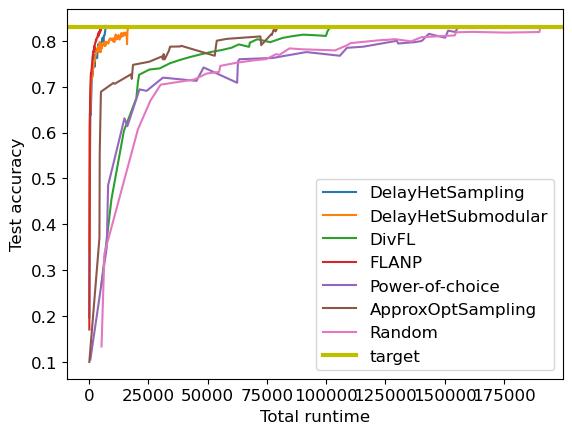}
\label{fig:fashionmnist_dir_nycmesh}}
\hfill
\subfloat[\normalsize Rotated CIFAR-10, NYCMesh delays]{\includegraphics[width=0.45\textwidth]{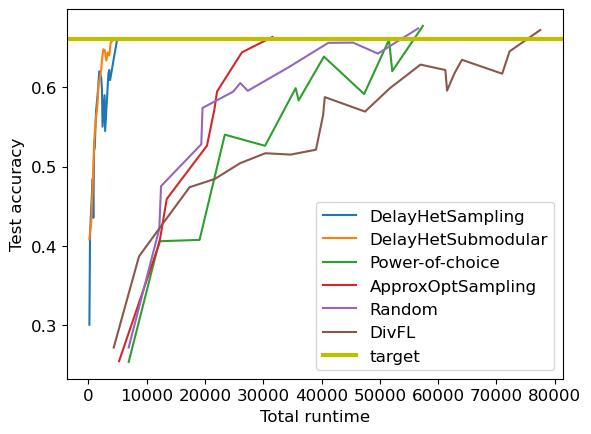}
\label{fig:cifar10_rot_nycmesh}}
\caption{\normalsize Test accuracies and total runtimes for all baselines on different dataset and delay distribution settings.}
\label{fig:plots}
\end{figure*}

\paragraph{Metrics}
We measure the wall-clock runtime for each algorithm until a target performance is achieved. For Quadratic dataset, this corresponds to test loss, while for other datasets, it is test accuracy. To compute total runtime, we add the maximum delay of clients selected in each round. From Theorems~\ref{thm:submodular} and ~\ref{thm:sampling}, our client selection schemes converge only to a neighborhood of the optima, therefore, our target loss/accuracies might slightly be worse than the best possible values for these datasets. We present the total runtimes for synthetic and NYCMesh delay distributions in  Tables~\ref{tab:uniform_runtime} and ~\ref{tab:nycmesh_runtime} respectively. We provide plots for total runtime vs. test accuracy for few dataset and delay combinations in Figure~\ref{fig:plots}. The evolution of test loss and total runtime for Quadratic dataset and NYCMesh delays is provided in Figure~\ref{fig:motivation} (c).
We required approximately $2$ week to run all our experiments with $3$ random seeds on a single Nvidia RTX 3090 GPU with $32$ GB RAM and $12$ CPUs. We provide code for running all our experiments in the supplementary.

\subsection{Results}

\paragraph{Quadratic} \texttt{Submodular} is the fastest algorithm followed by \texttt{Sampling}. DivFL and Power-of-Choice minimize heterogeneity, however, they still select slow clients. Therefore, they are always slower than our schemes for any test loss and slightly better than random. FLANP, which selects fastest clients, is faster than all baselines at the start. However, its later rounds make very little progress in terms of the test loss.

\paragraph{Simulated Datasets}
For these datasets, the level of heterogeneity is under our control. As we can see from the target accuracies, dirichlet label imbalance adds slightly more heterogeneity than rotations. For MNIST and FashionMNIST, even by rotations and dirichlet label imbalance, we are unable to add sufficient heterogeneity. Therefore, optimizing over only round delays by FLANP obtains best performance in a few cases. 
However, at least one of our schemes ranks second in terms of runtime, with both of our schemes surpassing all other baselines.
As CIFAR-10 is much more difficult to learn, rotations and dirichlet label imbalance do add sufficient heterogeneity. For CIFAR-10, it takes much longer to converge. Further, the convergence rate matters much more, making power-of-choice a competitive baseline to the best. As FLANP cannot adapt to the heterogeneity, we find that it fails to converge to the target accuracy for Rotated CIFAR-10.

\paragraph{Real Datasets}
Real datasets have higher data heterogeneity than Quadratic and simulated datasets. Our algorithms can adapt to this heterogeneity and  comprehensively outperform baselines. \texttt{Sampling} has a speedup of atleast $3.5\times$ over Random and at atleast $2\times$ over the nearest baseline.

\paragraph{Synthetic v/s NYCMesh delays}
Due to long-tail nature of NYCMesh, schemes which cannot accommodate delays (random, DivFL, Power-of-Choice) suffer. On the other hand, FLANP becomes a competitive baseline for datasets with low data heterogeneity, as delay becomes deciding factor for these cases. However, when data heterogeneity is also high (CIFAR-10, FEMNIST and Shakespeare), only our methods perform well with a significant speedup over baselines. This speedup is $10$ times for Rotated CIFAR-10 with NYCMesh delays. Both our methods are atleast $1.5$ times faster than the nearest baseline and atleast $20$ times faster than random.

\paragraph{Comparison to ApproxOptSampling~\cite{lat_het}}
From Table~\ref{tab:uniform_runtime}, with a simpler synthetic delay distribution, the performance of ApproxOptSampling is often worse than other baselines even Random (Quadratic, Dirichlet FashionMNIST, Dirichlet CIFAR10). This is due to the lack of warm-up steps. For long tail NYCMesh delays, from Table~\ref{tab:nycmesh_runtime}, ApproxOptSampling always outperforms random and is the next best method for all datasets except simpler ones (MNIST, FashionMNIST).  
Note that for FEMNIST, a real federated dataset with large data heterogeneity, ApproxOptSampling does not converge to the target accuracy. This is due to the poor modeling of heterogeneity in ApproxOptSampling. Further, our best performing methods for both synthetic and NYCMesh delays performs better than ApproxOptSampling. For NYCMesh delays, our methods are atleast $8x$ faster than ApproxOptSampling.

\paragraph{Justification for Feature Heterogeneity}
Note that DivFL uses a similar submodular formulation to our algorithm, however, it does not consider feature heterogeneity, but uses norm of the actual gradient difference. Throughout our experiements, even DivFL has not performed best or second best for any dataset. This shows that the connection between feature heterogeneity and speed is indeed what should be utilized.

\paragraph{Differences between \texttt{Submodular} and \texttt{Sampling}}
Both \texttt{Submodular} and \texttt{Sampling} compute feature heterogeneity $B_{ij}$ and then solve an optimization problem based on it. The only difference lies in the selection of a specific set of clients as compared to sampling from a probability distribution. Note that if our estimates for $B_{ij}$ and solve the optimization for both the algorithms fully, then \texttt{Sampling} would have slower convergence simply due to the presence of sampling variance. In practice, this does not happen, and \texttt{Submodular} can end up with a suboptimal set of clients. The randomness in \texttt{Sampling} scheme allows it to overcome this restriction. In general, for the combination of low data heterogeneity and low latency heterogeneity (MNIST and FashionMNIST with synthetic delays) \texttt{Sampling} seems to perform better, while for high data heterogeneity and high latency heterogeneity (Rotated heterogeneity or real datasets with NYCMesh delays), \texttt{Submodular} seems to perform better.

\subsection{The Impact of Various Heterogeneity}
To study the impact of various heterogeneity on our Algorithms~\ref{alg:submodular} and \ref{alg:optimalsampling}, we run our algorithms with different data and latency heterogeneity keeping all other parameters fixed.

\textbf{Data Heterogeneity} For data heterogeneity, we run Rotated MNIST with synthetic delays for different values of maximum rotation ($30,45,60$) and report the total runtime in Table~\ref{tab:add_exp} (Left). Increasing data heterogeneity increases the runtime to achieve same target for both algorithms. However, \texttt{DelayHetSampling} can handle high heterogeneity (increase of $85s$) much better than \texttt{DelayHetSubmodular} (increase of $546s$). This can be due to the ability of

\textbf{Latency Heterogeneity} For latency heterogeneity, we run Dirichlet MNIST with synthetic delays for different values of maximum network ($1,5$ and $10$ MB/s) and report the runtime in Table~\ref{tab:add_exp}. Increasing the max network speed decreases latency heterogeneity and rdecreases the runtime in all cases. Both algorithms appear to handle increase in delays equally well, as decreasing speed from $5$ MB/s to $1$ MB/s leads to only a $2$-fold increase in runtimes.
\begin{table*}[t!]
\normalsize
    \centering
    \begin{tabular}{ccc}
    \toprule
    \makecell{Maximum\\Rotation} & Submodular & Sampling\\
    \midrule
    $30^\circ$ & 619 & 293 \\
    $45^\circ$ & 845 & 513 \\
    $60^\circ$ & 1391 & 597 \\
    \bottomrule\\
    \end{tabular}
    \quad
    \begin{tabular}{ccc}
    \toprule
    \makecell{Maximum\\Link Speed} & Submodular & Sampling\\
    \midrule
    $1$ MB/s & 1362  &  447\\
    $5$ MB/s & 687 & 224 \\
    $10$ MB/s & 666 & 193 \\
    \bottomrule\\
    \end{tabular}
    \quad 
    \begin{tabular}{ccc}
    \toprule
    \makecell{Target\\Accuracy} & Submodular & Sampling\\
    \midrule
    $0.64$  & 522	 & 1018\\
    $0.67$  & 1101	& 1516 \\
    $0.7$  &3373	& 3246\\
    \bottomrule\\
    \end{tabular}\quad 
    \begin{tabular}{c}    \includegraphics[width=0.35\linewidth, height=4cm]{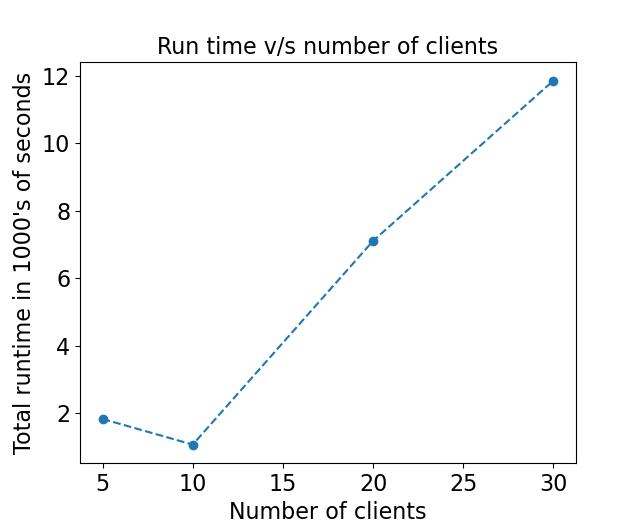}
    \end{tabular}\quad
    \begin{tabular}{ccccc}
    \toprule
    Dataset     &  \makecell{ 
    Feature\\  Heterogeneity} & Submodular & Sampling & ApproxOptSampling~\cite{lat_het}\\
    \midrule
    Quadratic 	& 195 & 0.68 & 105 & $31$\\
    Rotated CIFAR10 & 30 & 0.01	 & 42 & $4320$ \\
    Shakespeare & 128 & 0.03 & 90 & 718 \\    
    \bottomrule\\
    \end{tabular}
        \caption{\normalsize Top Left: Total runtime (in seconds) to achieve target accuracy for Rotated MNIST dataset with different values of maximum rotation. Top Right : Total runtime (in seconds) to achieve target accuracy for Dirichlet MNIST for different maximum link speed. Middle Left : Total runtime(in seconds) to achieve different values of Target accuracy for Dirichlet CIFAR-10. Middle Right: Total runtime to reach target MSE on Quadratic dataset with NYCMesh delay distribution for \texttt{DelayHetSampling} for different number of clients sampled per round. Bottom: Time spent(in seconds) calculating feature heterogeneity, performing submodular minimization, and computing the optimal sampling for different datasets in a single round.}
    \label{tab:add_exp}
\end{table*}

\subsection{Sensitivity Analyses}
\textbf{Number of Selected Clients} The set of clients selected by  \texttt{DelayHetSubmodular} is not under our control, but we can choose the number of clients sampled per round for \texttt{DelayHetSampling}. To assess the sensitivity wrt this parameter, we run \texttt{DelayHetSampling} on Quadratic dataset with NYCMesh delays for $5,10,20$ and $30$ clients selected per round. We plot the runtime till we reach the normalized target test loss of $2.95$ in Table~\ref{tab:add_exp}. If the number of clients is small, our algorithm cannot select enough clients to represent the heterogeneity in the dataset, while for a larger number of clients, the maximum delay of the selected set is often very large. Therefore, if the number of clients is very large or very small, the total runtime is high.

\textbf{Target Accuracy} 
In Table~\ref{tab:add_exp}(Right), we report the performance on Dirichlet CIFAR-10 with synthetic delay distribution for different target accuracies of $0.64, 0.67,0.7$. As is expected, increasing target accuracy increases the time required to achieve it. \texttt{DelayHetSubmodular} seems to perform worse than \texttt{DelayHetSampling} for higher target accuracies. As \texttt{DelayHetSubmodular} selects a fixed set, improper estimates of parameters or poor submodular minimization may lead to suboptimal client sets. From Theorems~\ref{thm:submodular}, the final accuracy is also determined by the selected set $\frac{\Gamma_S}{\mu(1-B_S)}$. For sampling, on the other hand, a selected sampling distribution might be suboptimal, but the randomness in sampling procedure makes it less restrictive.

\subsection{Runtime Overhead}
Both our client selection schemes perform two computationally intensive tasks per round -- computing feature heterogeneity for all pairs, and solving an optimization problem. For $m$ clients and $d$-dimensional features, computing features would require $\mathcal{O}(m)$ forward passes.  $\mathcal{O}(m^2 d^2)$ is required to pairwise feature heterogeneity from features as we need to find maximum eigenvalues. This step is also performed on a GPU. The submodular and sampling minimization algorithms take polynomial time in terms of problem parameters. We report the time taken for computing feature heterogeneity and performing submodular and sampling optimization in the first round for three different datasets with synthetic delays in Table~\ref{tab:add_exp}. We also report the corresponding time required for ApproxOptSampling~\cite{lat_het} to obtain it's sampling distribution in the same table as comparison. The overhead due to our algorithms is much smaller than the network delay, as our algorithms converge in ony a few rounds. Further, we find that submodular minimization is extremely fast while finding the optimal sampling requires atleast a minute. For Quadratic dataset, $m=100$ and $d=500$, so it is slowest. For CIFAR-10 $m=50$ is low and for Shakespeare $d=100$ is low. The dimension of features for NN models is the size of the last layer.
Pairwise heterogeneity computation can also be parallelized on the server to reduce its overhead.

\section{Conclusion}
Data and latency heterogeneity can substantially impede the convergence of FL, but previous client selection schemes addressing them either lack theoretical optimality or suffer from computational inefficiency during training. In this paper, we propose two novel theoretically optimal and computationally efficient client selection schemes, \texttt{Submodular} and \texttt{Sampling}, that can handle heterogeneity in both data and latency and obtain smaller total runtime than baselines in experiments. Extending our methods to  FL algorithms other than FedAvg, like SCAFFOLD~\cite{pmlr-v119-karimireddy20a}, is a promising direction for future work.

\section*{Acknowledgement}
This research was funded by TILOS (The Institute for Learning-Enabled Optimization at Scale, Grant Number : NSF 2112665). 

\bibliographystyle{plainnat}
\bibliography{references}

\appendix

\section{Proofs}
\label{sec:proofs}
\subsection{Proof of Lemma~\ref{lem:bias_grad_submod}}

\begin{align*}
    &\norm{\nabla f_S(w) - \nabla f(w)} = \norm{\sum_{i\in S} \alpha_i \nabla f_i(w) - \sum_{j=1}^m \nabla f_j(w)}\\
    &=\norm{\frac{1}{m} \sum_{j=1}^m ( \nabla f_{\beta_j(S)} (w) - \nabla f_j(w))}\\
    &\leq \frac{1}{m}\sum_{j=1}^m \norm{\nabla f_{\beta_j(S)}(w) - \nabla f_j(w)}\\
    &\leq \frac{1}{m}\sum_{j=1}^m (B_{j,\beta_j(S)} \norm{\nabla f(w)} + \Gamma_{j,\beta_j(S)})
\end{align*}
We use the value of coefficients from Definition~\ref{def:coeff}. Then, we apply the triangle inequality and use Assumption~\ref{assump:pair_het} to bound the difference in gradients. We collect the constant term and the terms of $\norm{\nabla f(w)}$ and square both sides to obtain the values of $B_S$ and $\Gamma_S$.

\subsection{Proof of Remark~\ref{rem:submodular}}
\label{sec:g_submod_proof}
For submodularity of a function $g:2^{[m]} \to \R$ is submodular, we need to show that for every set $\phi \subseteq S_1 \subset S_2 \subset [m]$ and $i\in [m]\setminus S_2$, we have $\Delta(i|S_1) \geq \Delta(i|S_2)$, where $\Delta(i|S) = g(S \cup \{i\}) - g(S)$ for any subset $S \subseteq [m]$.

To show this inequality for $g(S)$, we need to handle $B_S$ carefully.
First, we define a set function $h(S)= \frac{1}{m}\sum_{j=1}^m B_{j,\beta_S(j)}$ where $-h(S)$ is submodular. Note that $-h(S)$ is a facility location problem~\cite{pmlr-v119-mirzasoleiman20a,balakrishnan2022diverse}, therefore it is submodular. Further, $h(S)$ is monotonically decreasing.
Here, $B_S = 2h^2(S)$ and $\tau_S = \max_{i\in S}\tau_i$, then for any $i\in [m]\setminus S$, we have, 
% Therefore for any $S_1 \subset S_2 $ and $i \in [m]\setminus S_2$, we have
% \begin{equation}\label{eq:h_prop}
%          h(S_1\cup\{i\}) - h(S_1) \leq h(S_2\cup\{i\}) - h(S_2),\, 
%          h(S_1) \geq h(S_2)
% \end{equation}
% Now, we can express $B_S$ as $B_S = 2h^2(S)$. 

\begin{align*}
    \Delta(i|S) =& 
    \frac{2\tau_S(h^2(S\cup\{i\}) - h^2(S))}{(1 - 2h^2(S\cup\{i\}))(1 - 2h^2(S))}  \\
    &+ \frac{\max\{\tau_i -\tau_S,0\}}{1 - 2h^2(S\cup\{i\})}
\end{align*}

Now, to check if $g(S)$ is submodular, we need to see if $\Delta(i|S_1)\geq \Delta(i|S_2),\,\forall S_1 \subset S_2$. We end up with three cases based on delays $\tau_i,\tau_{S_1},\tau_{S_2}$.
Note that $\tau_{S_1} \leq \tau_{S_2}$. We obtain $3$ cases --

    \paragraph{Case I : $\tau_i \leq \tau_{S_1} \leq \tau_{S_2}$} : The second term in $\Delta(i|S_1)$ and $\Delta(i|S_2)$ is $0$. Then, using of $h(S)$, we obtain $h(S_1\cup\{i\}) + h(S_1) \geq h(S_2\cup\{i\}) + h(S_2),
        (1 - 2h^2(S_1\cup\{i\}))(1 - 2h^2(S_1)) \leq (1 - 2h^2(S_2\cup\{i\}))(1 - 2h^2(S_2))$
and $-(h(S_1\cup\{i\}) - h(S_1)) \geq - (h(S_2\cup\{i\}) - h(S_2))$.     Multiplying these equations with $-\tau_{S_1} \geq -\tau_{S_2}$, we obtain $\Delta(i|S_1) \geq \Delta(i|S_2)$.

    \paragraph{Case II : $\tau_{S_1} \leq \tau_i \leq \tau_{S_2}$} : For this case, the second term in $\Delta(i|S_2)$ is $0$ while it is positive for $\Delta(i|S_1)$. Since we have established that the first term of $\Delta(i|S_1)$ is $\geq$ the first term of $\Delta(i|S_2)$ adding a positive term will make it even larger. Therefore, in this case, $\Delta(i|S_1) \geq \Delta(i|S_2)$.
    
     \paragraph{Case III : $\tau_{S_1} \leq \tau_{S_2} \leq \tau_i$} :  In this case, both terms are present in both $\Delta(i|S_1)$ and $\Delta(i|S_2)$. Since from first case, we know that the first term is larger for $S_1$, we need to only analyse the second term. Since $\tau_i - \tau_{S_1} \geq \tau_i - \tau_{S_2}$ and $h^2(S_1\cup\{i\})\geq h^2(S_2\cup\{i\})$, even the second term is larger for $S_1$ making their sum larger. Therefore, $\Delta(i|S_1) \geq \Delta(i|S_2)$.

From the above three cases, we can conclude that $\Delta(i|S_1) \geq \Delta(i|S_2)$ and $g(S)$ is submodular.
% \section{Proofs for Section~\ref{sec:sampling}}
% \label{sec:proof_sampling}
\subsection{Proof of Lemma~\ref{lem:expected_bias}}
\label{sec:proof_sampling}
 We set $f_p(w) \triangleq \sum_{i=1}^m p_i f_i(w), \forall w\in \R^d$. Then, for any set $S\sim p^K$, we have,
\begin{align*}
     \E_S[\nabla f_S(w)] =\E_{i\sim p}[\nabla f_{i}(w)]= \sum_{i=1}^m p_i \nabla f_i(w) = \nabla f_p(w) 
\end{align*}

Now, to obtain the expected bias of $\nabla f_S(w)$, we first use the bias-variance decomposition for a random vector. If $X$ is a random vector, then $\E[\norm{X - a}^2] = \E[\norm{X - \E[X]}^2] + \norm{\E[X] - a}^2$ for any constant $a$.  Here, we set the value of $a$ to be $\nabla f(w)$ and $\nabla f_S(w)$ is the random variable.
We finally use Jensen's inequality and calculate the variance for $\nabla f_S(w)$.
\begin{align*}
    \E_{S}[\norm{\nabla f_S(w) - \nabla f(w)}^2]&= \E_{S}[\norm{\nabla f_S(w) - \nabla f_p(w)}^2] + \norm{\nabla f(w) - \nabla f_p(w)}^2\\
    &= \frac{1}{K}\sum_{i\in [m]} p_i \norm{\nabla f_i(w) - \nabla f_p(w)}^2  + \norm{ \frac{1}{m}\sum_{i\in [m]}(\nabla f_i(w) - \nabla f_p (w))}^2\\
    & \leq \sum_{i\in [m]}(\frac{p_i}{K} + \frac{1}{m})\norm{\nabla f_i(w) - \nabla f_p(w)}^2\\
    & \leq  \sum_{i\in [m]}(\frac{p_i}{K} + \frac{1}{m}) \norm{\sum_{j\neq i \in [m]} p_j(\nabla f_i(w) - \nabla f_j(w)}^2\\
    & \leq  \sum_{i\in [m]}\sum_{j\neq i \in [m]}(\frac{p_i}{K} + \frac{1}{m})p_j \norm{\nabla f_i(w) - \nabla f_j(w)}^2\\
    % & \leq  \sum_{i\in [m]}\sum_{j\neq i \in [m]}(\frac{p_i}{K} + \frac{1}{m})p_j \norm{\nabla f_i(w) - \nabla f_j(w)}^2\\
    &\leq \sum_{i\in [m]}\sum_{j\neq i \in [m]}2(\frac{p_i}{K} + \frac{1}{m})p_j  (B_{i,j}^2\norm{\nabla f(w)}^2 + \Gamma_{i,j}^2)
    % &\leq \left(\sum_{i\in [m]}\sum_{j\neq i \in [m]}2\left(\frac{p_i}{K} + \frac{1}{m})p_j B_{i,j}^2\right)\right)\norm{\nabla f(w)}^2 \\
    % &+ \left(\sum_{i\in [m]}\sum_{j\neq i \in [m]}2\left(\frac{p_i}{K} + \frac{1}{m}\right)p_j \Gamma_{i,j}^2\right)\\
\end{align*}
Here, we use bias-variance decomposition for any random variable $X$ and a deterministic constant $b$ independent of $X$, $\E_X[(X - b)^2] = \mathbb{V}ar(X) + (\E[X] - b)^2$. Further, if $X_1,X_2,\ldots, X_k$ are iid, then $\mathbb{V}ar(\frac{1}{K}\sum_{k=1}^K X_k) = \frac{1}{K}\mathbb{V}ar(X_1)$. Then, we use the Young's inequality for the second term. Finally, we use Assumption~\ref{assump:pair_het}. By collecting the terms of $\norm{\nabla f(w)}^2$ and constants and applying the definition of $B_p$ and $\Gamma_p$, we complete the proof.

\subsection{Proof of Proposition~\ref{lemma:pair_het_lin_reg}}
\label{sec:pair_het_lin_reg_proof}
For our linear regression example, $\nabla f_i(w) = A_i(w - w_i^\star)$ and $\nabla f(w) = A (w - w^\star)$. We try to fit this into Assumption~\ref{assump:pair_het}.
\begin{align*}
    \norm{\nabla f_i(w) - \nabla f_j(w)}  &=\norm{ (A_i - A_j) w - (A_i w_i^\star - A_j w_j^\star)}\\
    &=\lvert\lvert(A_i - A_j)A^{-1} A (w - w^\star)+ (A_i - A_j)A^{-1}w^\star + (A_i w_i^\star - A_j w_j^\star)\rvert\rvert\\
    &\leq \norm{ (A_i - A_j)A^{-1}}_2 \norm{\nabla f(w)} \\
    &\,+ \norm{(A_i - A_j)A^{-1}w^\star + (A_i w_i^\star - A_j w_j^\star)}_2
\end{align*}
We use triangle inequality and Cauchy-Schwarz.

\end{document}